# Deep Learning-based Sentiment Analysis in Persian Language

**Mohammad Heydari[a,*], Mohsen Khazeni[a], Mohammad Ali Soltanshahi[a]**

[a] Department of IT Engineering, Faculty of Industrial and Systems Engineering, Tarbiat Modares University,

[a]{m_heydari, m.khazeni, m.soltanshahi}@modares.ac.ir

**Abstract:** Recently, there has been a growing interest in the use of deep learning techniques for tasks in natural language processing (NLP), with sentiment analysis being one of the most challenging areas, particularly in the Persian language. The vast amounts of content generated by Persian users on thousands of websites, blogs, and social networks such as Telegram, Instagram, and Twitter present a rich resource of information. Deep learning techniques have become increasingly favored for extracting insights from this extensive pool of raw data, although they face several challenges. In this study, we introduced and implemented a hybrid deep learning-based model for sentiment analysis, using customer review data from the Digikala Online Retailer website. We employed a variety of deep learning networks and regularization techniques as classifiers. Ultimately, our hybrid approach yielded an impressive performance, achieving an F1 score of 78.3 across three sentiment categories: positive, negative, and neutral.

## Introduction

In the modern era, the surge in user access to a wide range of online communication platforms, including social networks and review and sales websites, has led to a massive increase in textual data production on these platforms. The analysis of this textual data is both practical and beneficial for various purposes. Nowadays, sentiment analysis finds applications across diverse sectors, including politics and business intelligence. One of the foremost online sales platforms in Iran is Digikala, where users post ratings and reviews on products from multiple perspectives after purchase, providing valuable feedback for future consumers. As such, Digikala has become one of the most significant sources of textual data for natural language processing and text mining in the Persian language.

## Background

As deep learning-based sentiment analysis emerges as a cutting-edge approach, there remains a limited body of research on the subject. Roshanfekr et al., presented a dataset of electronic product reviews in Persian and applied deep learning techniques to it [1]. Dashtipour and team developed a system combining Deep Auto Encoders and Deep CNNs for a novel dataset of Persian film comments, achieving an efficiency improvement with an accuracy of 82.86% over traditional Multi-Layer Perceptrons [2]. BokaeeNezhad and associates proposed a novel deep learning architecture for Persian sentiment analysis, utilizing CNNs for extracting local features and LSTMs for learning long-term dependencies, with Word2vec for unsupervised word representation, achieving an 85% accuracy rate [3]. Karimi and colleagues investigated the effectiveness of attention models and the BERT pre-trained language model in Persian due to the challenges of obtaining sufficient data for acceptable word representation vectors, showing superior performance over skip-gram, LSTM, and CNN models [4][5]. Zobeidi and team introduced a sentence-level opinion classifier leveraging Word2Vec and CNN for feature extraction and Bi-LSTM for sentiment classification, reaching a 95% accuracy rate [6]. Dashtipour et al. designed a novel hybrid system integrating Dependency Grammar-Based Rules with DNN for concept-level Persian Sentiment Analysis, using a Persian hotel reviews dataset and achieving 86.29% accuracy [7]. Rohanian et al. proposed a method for classifying Persian sentences into two and five categories using a convolution layer over input text data, demonstrating the advantages of CNN over traditional machine learning algorithms, especially for small text data, validated across three datasets with the AUC metric [8]. Dastgheib and colleagues introduced a hybrid approach combining SCL and CNN, where the SCL method selects the most useful pivot attributes for domain adaptation without efficiency loss, improving sentiment classification in two domains by over 10%. This summary highlights recent advancements and studies in the field of deep learning-based sentiment analysis for Persian texts.

## Dataset

The dataset utilized in this research is segmented into 14 primary categories. Additionally, the data are categorized into various class categories, such as datasets with positive and negative classes. For instance, we have datasets divided into three classes: positive, neutral, and negative. This research dataset is provided by the Digikala Online Retailer Open Data Mining Program, comprising 100,000 customer reviews across different product categories. All reviews are in the Persian language. The dataset includes the following fields: Product ID, Product Title, Title in English, User ID, Likes, Dislikes, Verification Status, Recommendation Status, Review Title, Comment, Pros, and Cons. A bar chart illustrating the distribution of data grouped by text length is presented in Figure 1.

## A. Normalization

We process the user comments file by eliminating unnecessary columns and converting the labels into binary form, represented by 0 and 1. Subsequently, we standardize the text formatting through several key steps:

1. removing punctuation,
2. eliminating letters and numbers, and,
3. filtering out stop words.

Regular expressions (regex) are employed to efficiently perform all these tasks, ensuring a clean and standardized dataset ready for further analysis.

## B. Tokenization

After the text normalization process, the next step involves transforming sentences into vectors, with each word being converted into a token. This tokenization phase is initiated using the Natural Language Toolkit (NLTK) library, which provides the necessary tools to break down the text into individual elements, facilitating the conversion of text into a format that can be analyzed or used in machine learning models.

## C. Sentence Length Unification

At this stage, we analyze the length of the sentences by counting the number of tokens they contain. We observe that some sentences are shorter than five words, while others exceed twenty words. Our analysis reveals that 90% of the sentences contain fifteen words or fewer. Based on this finding, we decide to set a threshold of fifteen words. For sentences that are longer than this threshold, we will select only the first fifteen words to ensure consistency and manageability in our dataset processing.

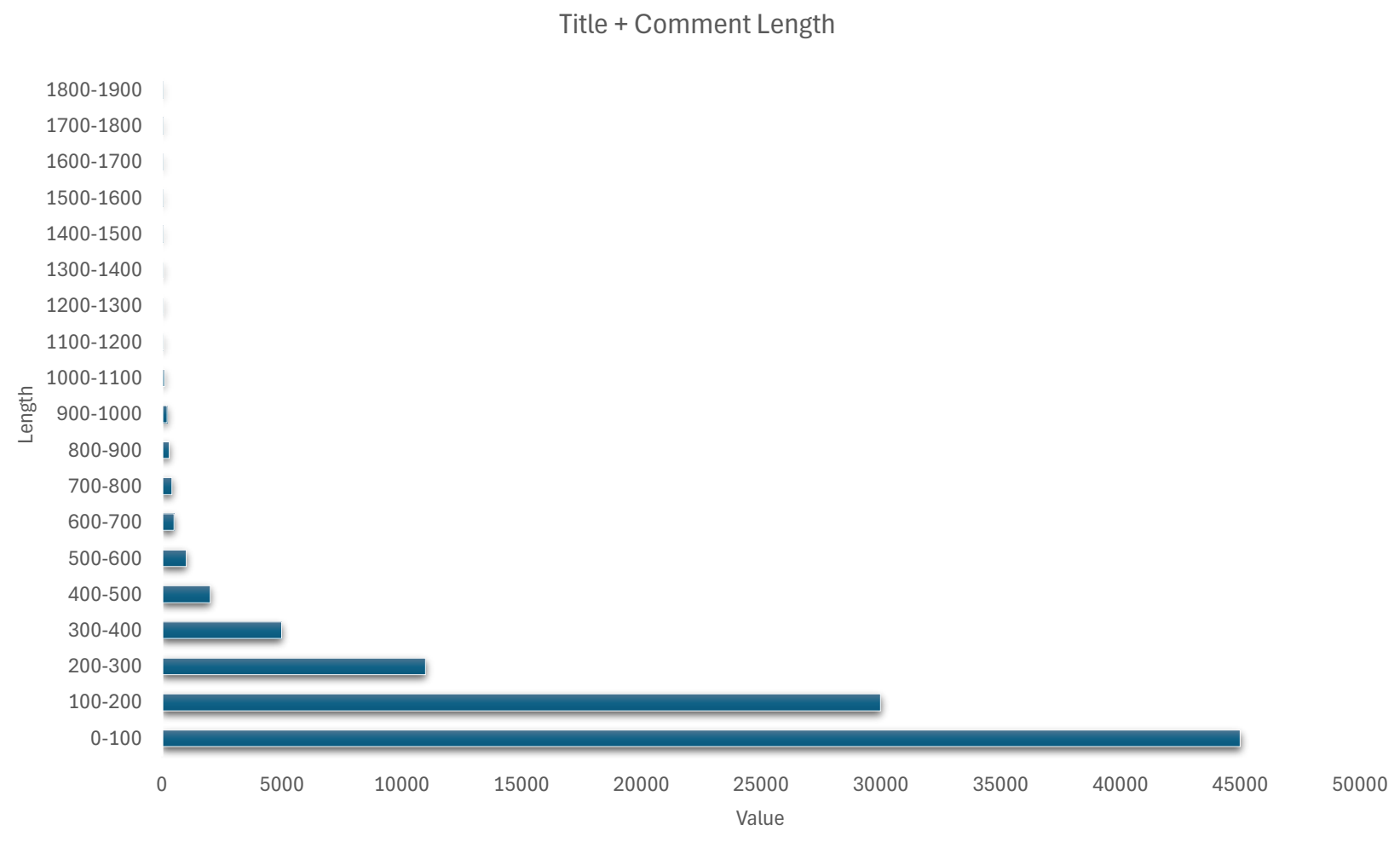

Figure 1 - Data Group by Text Length

## D. Vectorization

There are several methods for converting words into vectors, such as Term Frequency (TF), TF-IDF, among others [9][10]. A particularly renowned approach for capturing semantic relationships between words was introduced by Mikolov in 2013 [11][12]. It's clear that the larger the dataset, the better the semantic relationships between words can be captured. This is a key reason for utilizing pre-trained datasets. One of the well-known pre-trained datasets is GloVe, which is trained on 6 billion tokens and includes over 400,000 different words [13]. It is available in various dimensions including 50, 100, 200, and 300. For our purposes, we use the pre-trained GloVe model to convert words into 50-dimensional vectors. It's important to note that some words may not be found in the model due to incorrect typing, and in such cases, they are assigned a vector of zeros.

## E. Splitting of Train and Test Data

The dataset comprises users' comments on various products from the Digikala online retailer website. After randomly shuffling all the sentences, we allocate 80% of the dataset for training purposes and reserve the remaining 20% for testing. Given the substantial size and complexity of the data, we encountered limitations with available RAM capacity. To manage this, we divided the data into batches of 200,000 and stored them on disk. During the training phase, we sequentially load these batches into memory for processing. This approach allows for efficient handling of the data while mitigating memory constraints.

## Network Architecture

In this section, we explore various architectures and techniques for training neural networks to attain optimal accuracy. The focus begins with the simplest type of network, which lacks any hidden layers and employs a SoftMax function in the output layer. This configuration is often referred to as a logistic regression model when applied to classification tasks, despite being a neural network due to its structure. The absence of hidden layers means that this model directly maps input features to the output classes through a single transformation layer. Such a network is particularly useful for understanding the fundamental mechanics of neural operations and serves as a baseline for comparing more complex architectures.

*A. Network with non-hidden layer and SoftMax as output layer*

The inputs are the vectors of the sentences. These inputs can also be assigned as batches to the network. The size of the input dimension is as (1):

$$Batch\ Size\ *\ 15\ *\ 50$$

Indicates 15 words per sentence and 50 dimensions per word. The output dimension is 2. It means the output is either 01 or 10.

*B. Loss Function*

The Loss calculation function is also an entropy function that is calculate as (2):

$$cross_entropy\ =\ -tf.reduce_mean(Y_\ *\ tf.log(Y))$$

$Y_$ is the main output and $Y_$ is the activity function. The mean multiplication of this value is equal to the error as (3):

$$H(p,q) = -\sum_{\{x\ \in X\}p(x)\backslash log} q(x)$$

According to studies, this error function works better than other functions such as MSE.

*C. Activation Function*

The activity function is calculated as follows. When the input vector multiplied by weights and multiplied by bias as (4):

$$Y\ =\ tf.nn.softmax(tf.matmul(XX,W)\ +\ b)$$

In addition, this value passed to the SofMax function as:

$$\mathrm{Softmax}(L_n) = \frac{e^{L_n}}{|e^L|}$$

*D. Optimization Algorithm and Learning*

The technique applied in this scenario is gradient descent, with a set learning rate of 0.003.

## Results

In this segment, we describe the models employed in our research along with their ultimate data metrics. The extensive and complex nature of the data necessitated a powerful computing infrastructure. Our system was equipped with:

An Intel Core i7 8750 CPU with a 4 GHz clock speed, 2) 16 GB DDR4 RAM, 3) An M2 SSD as the primary storage device, and 4) A GeForce GTX 1060 GPU. The Network's architecture data are presented in Table 1, the Performance of the Models on the Digikala Dataset in Table 2 and the User Comment Statistics in Table 3.

*A. Word2Vec with a SoftMax Layer*

The network's output after training for 100 epochs is depicted. The highest accuracy recorded was 63.1, achieved at epoch 13, indicating that the network learns at a slow pace and exhibits suboptimal learning capability.

*B. Multi-Layer Perceptron (MLP) with Five Hidden Layers and Sigmoid Activation*

We enhanced the initial network by incorporating five hidden layers with a sigmoid activation function. The error optimization is managed by the Adam function. The network features 200, 100, 60, and 30 hidden neurons, respectively. The output layer uses a SoftMax activation function, while all other parameters remain consistent with the first network. The learning rate begins to decay if there is no accuracy improvement after 27 epochs. Beyond this point, the network's accuracy showed an uptick, signifying an improvement over the previous model, although overfitting remains a concern.

*C. MLP Network with Five Hidden Layers, ReLU Activation, and Learning Rate Decay*

This model retains the same parameters as the previous one but incorporates the ReLU activation function and a dynamic learning rate that adjusts depending on the epoch number. This modification enhances the network's ability to avoid local minima. The learning rate is calculated using the following equation (6):

$$0.0001 + 0.003 * (1/e)^\wedge(step/2000))$$

Table 1 - Network's Architecture Data

| Attribute | Value |
|---|---|
| **Batch Size** | 512 |
| **Learning Rate** | 0.001 |
| **Learning Rate Decay** | 0.9 |
| **Drop out** | 0.5 |
| **Number of LSTM cell for each word** | 100 |
| **Number of LSTM cell for each Character** | 50 |

Table 2 - Performance of Models on Digikala Dataset

| Model | F1 |
|---|---|
| **Word2Vec+SoftMax** | 63.1 |
| **Word2Vec+5Layers+Sigmoid** | 72 |
| **Word2Vec+5Layers+ReLU+lrDecay** | 72.1 |
| **Word2Vec+5Layers+ReLU+lrDecay+DropOut** | 71.8 |
| **Word2Vec+LSTM** | 75.7 |
| **CharEmbed+Word2Vec+LSTM+RUS** | 73.9 |
| **CharEmbed+Word2Vec+LSTM** | 78.3 |

Table 3 - User's Comments Statistics

| Category | Positive | Negative | Neutral |
|---|---|---|---|
| **IT (General)** | 2673 | 803 | 537 |
| **Digital Products** | 6435 | 3885 | 1768 |
| **Home Appliances** | 4183 | 1922 | 1446 |
| **Mobile** | 546 | 107 | 92 |
| **Beauty, Health** | 4142 | 1803 | 1360 |
| **Personal Assistant** | 2243 | 824 | 504 |
| **Cars, Tools, Office Supplies** | 3687 | 1513 | 1030 |
| **Kitchen Appliances** | 3070 | 1369 | 927 |
| **Books, Stationery and Art** | 2171 | 746 | 561 |
| **Electric (Home)** | 1543 | 439 | 365 |
| **Audiovisual Equipment** | 1167 | 418 | 252 |
| **Fashion and Apparel** | 4057 | 1831 | 1408 |
| **Health, Detergents, Perfumes** | 399 | 142 | 96 |
| **Baby & Child Health** | 166 | 69 | 39 |

In this research, the term 'step' denotes the number of epochs completed, with the learning rate decreasing from an initial 0.0031 to a lower bound of 0.0001. Upon completion of 5 epochs, the model attained a 72.01% accuracy rate, reflecting a modest enhancement over the preceding model. Nonetheless, the issue of overfitting persists as a significant obstacle.

*D. MLP Network with 5 Hidden Layers, ReLU Activation Function, LR Decay and Dropout technique*

The current network retains identical parameters to the previous one, with the addition of a dropout strategy that eliminates 25% of the neurons along with their connected edges in each hidden layer. After five epochs, the network achieved an accuracy of 72.01%, which is a slight decrease from the former model. However, the notable improvement is the successful mitigation of network overfitting, attributed to the integration of the dropout technique.

*E. Long Short-Term Memory Network (LSTM)*

In this iteration of the network, the hidden layers from the earlier architecture have been replaced with an LSTM layer that houses 100 hidden neurons. After undergoing 9 epochs, the network's accuracy escalated to 75.07%, marking a significant improvement. This enhancement can be credited to the inherent design of recurrent networks, like LSTM, which consider the sequence and positional significance of words within a sentence.

*F. Bidirectional Recurrent Neural Networks – BiLSTM*

After 14 epochs the network reached 78.03 which is almost equal to the precision of the previous network.

G. Gated Recurrent Unit – GRU Network

The GRU network has a simpler structure than the two previous networks, so we can see that the network output is less accurate. After 6 epochs the network reached an accuracy of 73.09 which is less accurate than the LSTM network family.

H. Most important causes of people's satisfaction and dissatisfaction about products:

Ngrams: The most fundamental language model is the n-gram. In the fields of computational linguistics and probability, ngram is a continuous sequence of N words. The main idea behind the n-gram language model is to collect statistical information about the frequency of different n-grams. The result can enhance the forecasting of QAC . Ngrams visualization is presented in Fig. 2, Fig. 3, Fig. 4, and Fig. 5.

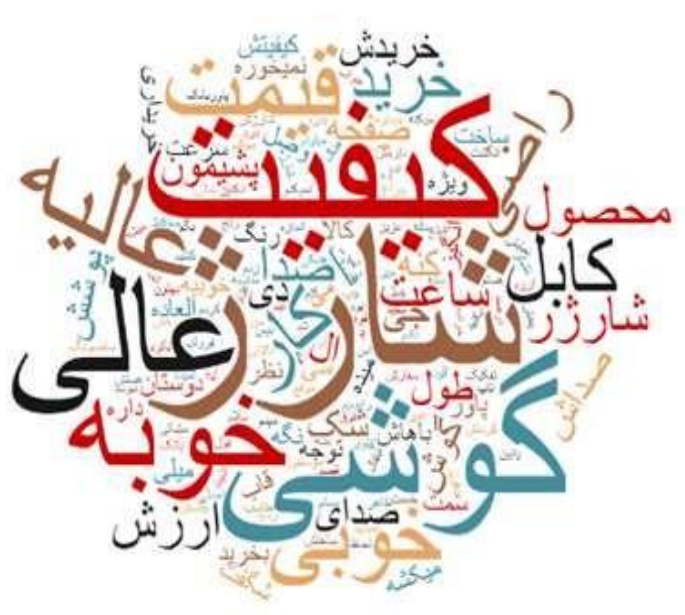

Figure 2 - Users satisfaction reason with digital goods category.

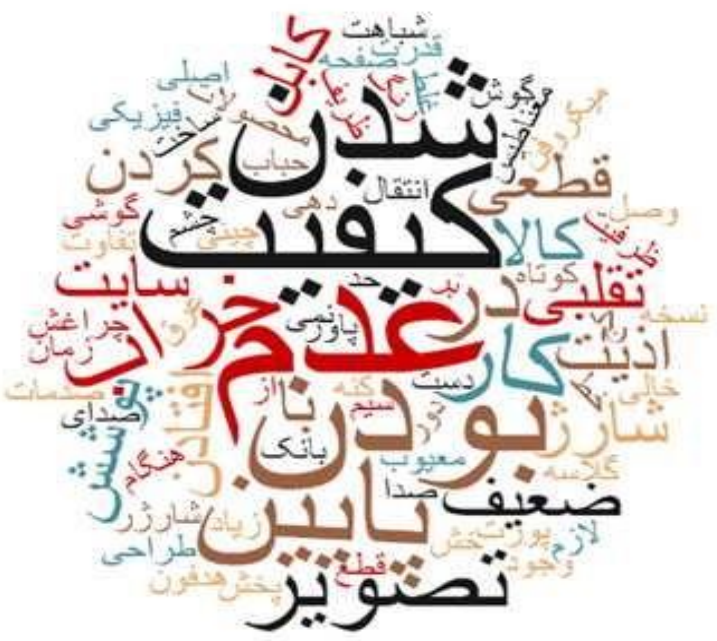

Figure 3 - Users dissatisfaction reasons with digital goods category.

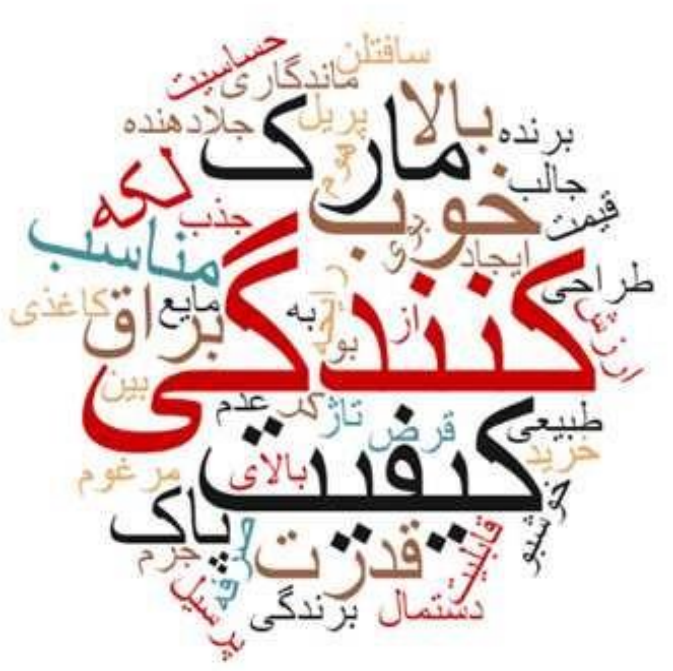

Figure 4 - Users satisfaction reasons with Home and kitchen goods category.

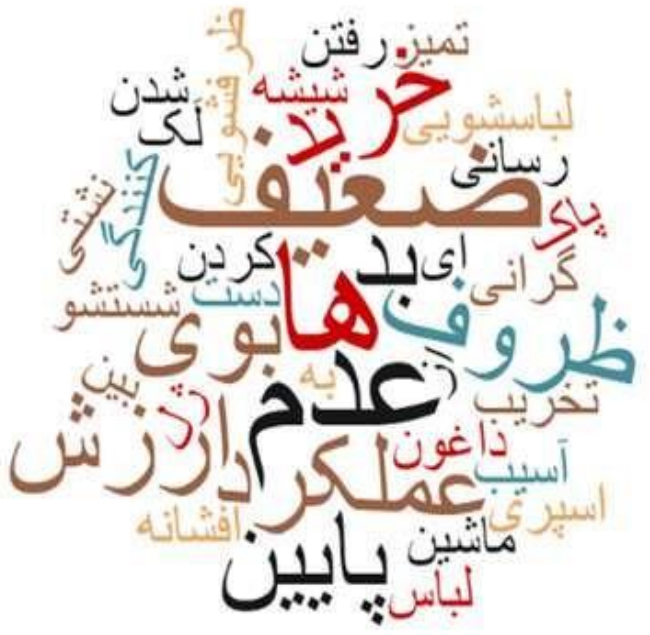

Figure 5 - Users dissatisfaction reason with the Home and kitchen goods category.

## Conclusion

This study introduces a novel hybrid deep learning model for sentiment analysis, which was tested on a dataset of customer reviews from the Digikala Online Retailer, a Persian-language website. A distinctive feature of this research is the model's application to three categories of sentiments—positive, negative, and neutral—which is a departure from many studies that limit their analysis to only positive and negative sentiments.

Our investigation faced two primary obstacles that challenge the enhancement of accuracy:

1. The scarcity of expansive Persian language training datasets and
2. The requirement for high-performing GPUs.

The driving force behind our research is to leverage deep learning methods and diverse models specifically tailored for the Persian language. Sentiment analysis is essential for Persian, encompassing its unique conversational and formal nuances. Furthermore, the potential for improved accuracy through hybrid methodologies is highlighted in recent literature, such as integrating character embeddings with word embeddings. Considering the burgeoning interest in applying cutting-edge deep learning methods to natural language processing, our future endeavors will explore the application of advanced transformer-based techniques to various Persian datasets.

## References

[1] B. Roshanfekr, S. Khadivi, and M. Rahmati, “Sentiment analysis using deep learning on Persian texts,” in *2017 Iranian Conference on Electrical Engineering (ICEE)*, May 2017, pp. 1503–1508, doi: 10.1109/IranianCEE.2017.7985281.

[2] K. Dashtipour, M. Gogate, A. Adeel, C. Ieracitano, H. Larijani, and A. Hussain, “Exploiting Deep Learning for Persian Sentiment Analysis,” *Lect. Notes Comput. Sci. (including Subser. Lect. Notes Artif. Intell. Lect. Notes Bioinformatics)*, vol. 10989 LNAI, pp. 597–604, Aug. 2018.

[3] Z. B. Nezhad and M. A. Deihimi, “A COMBINED DEEP LEARNING MODEL FOR PERSIAN SENTIMENT ANALYSIS,” *IIUM Eng. J.*, vol. 20, pp. 129–139, 2019.

[4] S. Karimi and F. Sadat Shahrabadi, “Sentiment analysis using BERT (pre-training language representations) and Deep Learning on Persian texts,” 2019.

[5] A. L. Maas, R. E. Daly, P. T. Pham, D. Huang, A. Y. Ng, and C. Potts, “Learning Word Vectors for Sentiment Analysis,” 2011.

[6] S. Zobeidi, M. Naderan, and S. E. Alavi, “Opinion mining in Persian language using a hybrid feature extraction approach based on convolutional neural network,” *Multimed. Tools Appl.*, vol. 78, no. 22, pp. 32357–32378, Nov. 2019, doi: 10.1007/s11042-019-07993-4.

[7] M. Rohanian, M. Salehi, A. Darzi, and V. Ranjbar, “Convolutional Neural Networks for Sentiment Analysis in Persian Social Media,” Feb. 2020.

[8] M. Dastgheib, S. Koleini, F. R.-I. J. of Information, and undefined 2020, “The application of Deep Learning in Persian Documents Sentiment Analysis,” *5.190.58.17*.

[9] J. Ramos and others, “Using tf-idf to determine word relevance in document queries,” in *Proceedings of the first instructional conference on machine learning*, 2003, vol. 242, no. 1, pp. 29–48.

[10] S. Robertson, “Understanding inverse document frequency: on theoretical arguments for IDF,” *J. Doc.*, 2004.

[11] T. Mikolov, I. Sutskever, K. Chen, G. Corrado, and J. Dean, “Distributed Representations of Words and Phrases and Their Compositionality,” in *Proceedings of the 26th International Conference on Neural Information Processing Systems - Volume 2*, 2013, pp. 3111–3119.

[12] T. Mikolov, K. Chen, G. Corrado, and J. Dean, “Efficient estimation of word representations in vector space,” *arXiv Prepr. arXiv1301.3781*, 2013.

[13] J. Pennington, R. Socher, and C. Manning, “GloVe: Global Vectors for Word Representation,” in *Proceedings of the 2014 Conference on Empirical Methods in Natural Language Processing ({EMNLP})*, Oct. 2014, pp. 1532–1543, doi: 10.3115/v1/D14-1162.

## Appendix.

Table 4 - Sample User’s Comments About Different Products Categories

| Category | User’s Comments |
|---|---|
| **IT** | واقعا عالیه، من که ازش خیلی راضیم. |
| **Home Appliance** | گیره‌های فلزی پولیشر خیلی سخت تا می‌شوند و لذا حوله خیلی سخت در می‌آید. |
| **Mobile** | گوشی این قیمت و امکانات در مقایسه با سایر برندها هیچ مزیتی نداره. |
| **Trimming Machine** | این ماشین قیمت فوق العاده رقابتی داره، آدمو ترغیب می‌کنه برای خرید این محصول. نکته‌ی دیگه، من با این قیمت انتظار نداشتم اصلاح به این منظمی رو داشته باشه. از سیستم کارکردش هم از طریق باتری و هم مستقیماً از طریق برق بود. با این رده قیمتی عالی می‌شد. |
| **Player** | عضوی از تیم گفتن این بخش ضعیف‌تر از نسخه‌های سابق خورده است اما با تحقیقاتی که داشتم، گفتن تفاوتی نداره و تفاوتش در سابق خوردن و تعداد خروجی‌هاست. اگر به سیستم موردنظر خودتون و تعداد خروجی زیاد براتون مهم نیست، تو این رنج قیمت بهترین گزینه هست. |